\documentclass[conference]{IEEEtran}
\usepackage[utf8]{inputenc} 
\IEEEoverridecommandlockouts
\usepackage{cite}
\usepackage{amsmath,amssymb,amsfonts}
\usepackage{algorithmic}
\usepackage{graphicx}
\usepackage{hyperref}
\usepackage{textcomp}
\usepackage{xcolor}
\def\BibTeX{{\rm B\kern-.05em{\sc i\kern-.025em b}\kern-.08em
    T\kern-.1667em\lower.7ex\hbox{E}\kern-.125emX}}
\begin{document}

\title{The Role of Modularity and Neuro-Regulation for the Production of Multiple Behaviors}

\author{\IEEEauthorblockN{Victor Massagué Respall}
\IEEEauthorblockA{\textit{Institute of Robotics} \\
\textit{Innopolis University}\\
Innopolis, Russia \\
v.respall@innopolis.ru}
}

\maketitle

\section{Introduction}
Nowadays, a long term goal for the robotics and artificial intelligence community is creating agents with the ability to solve different problems, producing multiple behaviors \cite{10.1371/journal.pcbi.1004128, duan2016benchmarking} . This task is challenging because until now, in order to learn a new behavior to solve another problem, previous acquired knowledge is forgotten to accommodate the new. This happens due to the fact that the connections of the neural network change when learning to perform a new behavior or task, leading to the loss of previously trained experience. 

This project investigates whether functional specialization or modularity can support the development of multiple behaviors. With the term functional specialization we refer to a situation in which one or more neurons, eventually forming a specific sub-part of the neural network policy, are primarily responsible for the production of a specific behavior and are less involved in the production of other behaviors. When the groups of specialized neurons are more connected with the neuron of the group than with other neurons, the groups of specialized neuron form a specialized neural module. 

In principle, modular solutions of this type can facilitate the development of multiple behaviors since each module is responsible for the production of a different behavior. Consequently, the interfaces that arise neural mechanisms supporting the production of different behaviors can be reduced. 

The realization of a structural modularity of this type, however, also require the availability of regulatory mechanisms that enhance the activity of the neurons specialized for the production of the behavior that is relevant in the current context and suppress or filter-out the effect of the neurons specialized for the production of alternative behaviors. 

The project involves the implementation of regulatory networks of this type and the realization of experiments involving the production of different behaviors. 

The paper is organized as follows: Section 2 describes the system setup for the experiments carried out explained in Section 3. Finally Section 4 briefly discusses the results. 

\section{Experimental Setup}
In this section the methodology followed to solve the task is described. For this project, evorobotpy \cite{nolfi2010evorobot} software library was used to obtain different environments and also train the robots in an evolutionary manner to produce multiple behaviors with the Salimans algorithm \cite{salimans2017evolution}. The first step to carry on this project consists in choosing a problem and designing a reward that encourage the development of multiple behaviours. Two environments of the library were used to evaluate the performance of the neuro-regulation method with different circumstances.

The next step, consists in implementing a neural network including regulatory neurons. Regulation can be realized by using special neurons that control the desired impact of other neurons. For example, by including in the policy network standard neurons and logistic regulatory neurons and by multiplying the output of standard neurons for the activation of the associated regulatory neurons.

To test whether neuro-regulation can improve the efficacy of the model evonet library is modified to adapt such behavior. More specifically, the layer of internal neurons is divided in two groups. 
\begin{itemize}
    \item The first group which contains the first half of the internal neurons is updated with the tanh activation function
    \item The second half of internal neurons is updated with the logistic function that returns a value between 0.0 and 1.0.
    \item The activation of the first half is multiplied with the activation of the second half of internal neurons. 
    \item The activation of the second half of internal neurons should be then set to 0.0. This implies that the connections weights from the second half of hidden units to the motor neurons will be useless, meaning that they will not have any impact on the output neurons.
\end{itemize}

This would imply that some of the internal neuron might be used more during the production of one behavior and some internal neurons can be used more during the production of the second behavior. In other words, neurons can specialise for the production of different behaviors, when necessary. The second set of internal neurons is used to perform the regulatory function. This is a possible way to allow a specialisation of the neurons.

\subsection{Hopper}
\begin{figure}
    \centering
    \includegraphics[width=\linewidth]{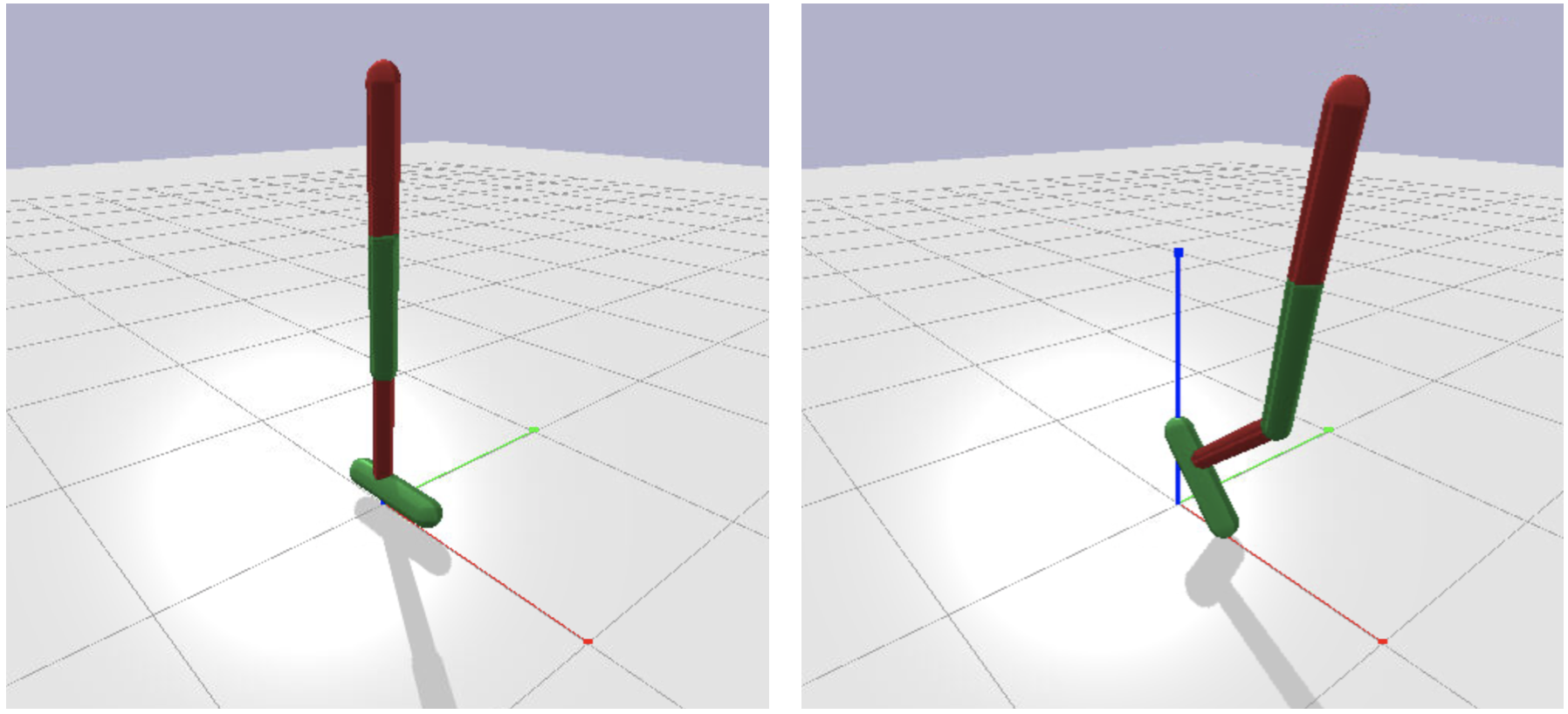}
    \caption{The Hopper robot environment, in rest (left) and in motion (right)}
    \label{fig:hopper}
\end{figure}

For the first set of experiments, a simple problem was selected, Hopper (see Figure \ref{fig:hopper}), to obtain many replication results fast to compare statistics across replications using different methodologies. The robot has three actuators, thigh, leg and foot joints, which is why it is a pretty simple problem to work with, because of its small number of actions.  

The goal for this robot is to learn two differentiated behaviors, one which tries to maximize the forward speed (by jumping forward) and the other which maximizes the vertical height (by jumping up).

The original reward function computes the distance covered toward the target destination during a step, which is already one of the behaviors that we want. Equation \ref{eq:beh1} shows how the reward for the first behavior is calculated, 

\begin{equation}
\label{eq:beh1}
    reward_{bh1} = \frac{d_{old}-d}{\Delta t}
\end{equation}
where $d$ represents the current distance from the robot to the goal and $d_{old}$ the previous distance to the goal. 

A second reward function (see Equation \ref{eq:beh2}) is added to represent the behavior of jumping vertically, on the basis of the elevation of the hopper over the floor and punish the robot slightly for moving forward. The punishment can be helpful to push the robot to produce differentiated behaviours but should not be too strong. Moreover, the second reward is weighted with a constant so that the maximum value of an effective robot can gain by jumping vertically is similar to the maximum value that an effective robot can gain by jumping forward. This is done to avoid one behavior achieving a much stronger or much weaker reward than the other behavior.

\begin{equation}
\label{eq:beh2_2}
    progress\_up = \left|\frac{h-h_{old}}{\Delta t}\right|
\end{equation}
where, $h$ is the current height above the ground of the Hopper and $h_{old}$ is the previous height above the ground.
\begin{equation}
\label{eq:beh2}
    reward_{bh2} = 2.0*progress\_up-0.5*reward_{bh1}
\end{equation}
where $progress\_up$ is described in Equation \ref{eq:beh2_2} and $reward\_{bh1}$ corresponds to the penalty for moving forward calculated as Equation \ref{eq:beh1}.

Next, a mechanism is needed to tell the robot which behavior (jumping forward or vertically) should perform. To do so, the observation vector was extended with two additional input neurons, called \textit{Behavior 1} and \textit{Behavior 2}, that are set to 5.0 and 0.0 when the robot should jump forward and to 0.0 and 5.0 when the robot should jump vertically. The observation vector which is the input to the neural network is shown in Table \ref{tab:obs}. The output of the neural network (motor neurons) has size 3 and controls the torque applied to each joint.

\begin{table}[!h]
\centering
\caption{Observation Vector of the Hopper robot (Input Neurons)}
\begin{tabular}{|l|l|}
\hline
0 - Height above ground & 1 - sin(angle target) \\
\hline
2 - cos(angle target)  & 3 - Velocity in x  \\
\hline
4 - Velocity in y   & 5 - Velocity in z    \\
\hline
6 - Roll angle          & 7 - Pitch angle  \\
\hline
\multicolumn{2}{|l|}{8,..,13 - Thigh, Leg, and Foot joints pos. and vel.}\\
\hline
14 - Feet contact        & 15 - Behavior 1    \\
\hline
16 - Behavior 2          &      \\
\hline
\end{tabular}
\label{tab:obs}
\end{table}

The parameters used for the experiments are a feed-forward neural network structure with one hidden layer of 50 neurons. The maximum duration of each episode was set to 500 steps with the termination condition of the robot falling down. Regarding the parameters of the algorithm, a total 50 million evaluation steps were used, and 11 replications to compare statistically the results obtained.

\subsection{Ant}
\begin{figure}
    \centering
    \includegraphics[width=0.5\linewidth]{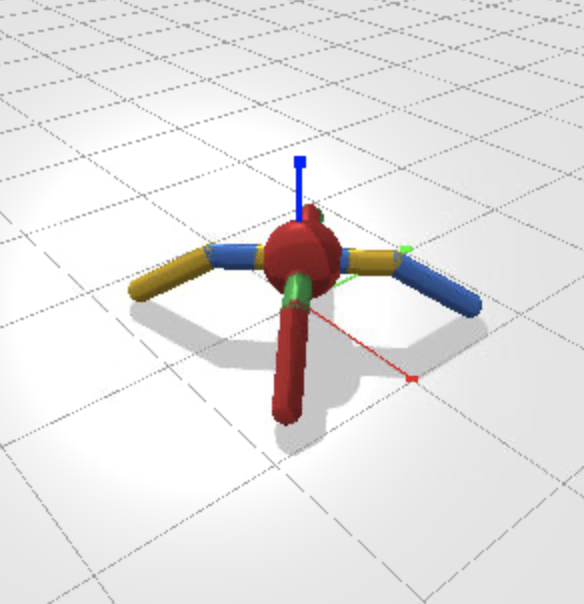}
    \caption{The Ant robot environment in rest}
    \label{fig:ant}
\end{figure}

This environment presents a more challenging and complex task with respect to the previous one. In this case the robot is the Ant (see Figure \ref{fig:ant}), which has 8 actuators, one hip and ankle joint for each of the four legs. This implies that the possible number of actions to take is much bigger, thus making the problem more difficult. 

The goal of this robot is to learn two differentiated behaviors, one which tries to maximize the distance walked $45^\circ$ left \ref{eq:bh1} with respect to the orientation of the robot and another behavior which tries to maximize the distance walked $45^\circ$ right \ref{eq:bh2}. 

The following equations describe the rewards for each behavior:

\begin{align}
    &xy\_angle = atan2(y-y_{old}, x-x_{old}) \label{eq:xy-angle}\\
    &self\_angle = xy\_angle - yaw \label{eq:self-angle}\\
    &step\_length = \sqrt{(x-x_{old})^2+(y-y_{old})^2} \label{eq:step}\\
    &reward_{bh1} = step\_length*\cos{(self\_angle-\frac{\pi}{4})} \label{eq:bh1}\\
    &reward_{bh2} = step\_length*\cos{(self\_angle+\frac{\pi}{4})} \label{eq:bh2}
\end{align}

Equation \ref{eq:xy-angle} calculates the angle of the vector composed by the previous and the current $xy$ position with respect to the x-axis, where $x, x_{old}, y, y_{old}$ correspond to the current $x$ and previous $x$ position of the robot, and current $y$ and previous $y$ position of the robot, respectively. Equation \ref{eq:self-angle} calculates the difference between the current orientation of the robot, $yaw$, and the angle walked. Next, Equation \ref{eq:step} gets the distance travelled for that period of time. Finally, for the first behavior, Equation \ref{eq:bh1} use the length walked multiplied by the cosine of the difference between $self\_angle$ and $45^\circ$, which is maximized when the robot walks exactly $45^\circ$ left and the second behavior, Equation \ref{eq:bh2}, where instead of the difference is the sum to be maximized when the robot walks exactly $-45^\circ$.

Following the same fashion as the Hopper robot, the observation vector was extended with two additional neurons, \textit{Behavior 1} and \textit{Behavior 2}, to represent both behaviors. When they are set to $5.0$ and $0.0$ respectively, the robot is rewarded for walking $45^\circ$ left, and when they are set to $0.0$ and $5.0$ the robot should walk right $45^\circ$. Thus, the resulting observation vector is shown in Table \ref{tab:obs_ant}, corresponding to the 30 input neurons. The output of the neural network has size 8 and controls the torque applied to each joint. 

\begin{table}[!h]
\centering
\caption{Observation Vector of the Ant robot (Input Neurons)}
\begin{tabular}{|l|l|}
\hline
0 - Height above ground & 1 - sin(angle target) \\
\hline
2 - cos(angle target)  & 3 - Velocity in x  \\
\hline
4 - Velocity in y   & 5 - Velocity in z    \\
\hline
6 - Roll angle          & 7 - Pitch angle  \\
\hline
\multicolumn{2}{|l|}{8,..,23 - Hip and Ankle joints pos. and vel. for each of the four legs}\\
\hline
\multicolumn{2}{|l|}{24,..,27 - Feet contact for each of the four legs}\\
\hline
28 - Behavior 1  & 29 - Behavior 2  \\
\hline
\end{tabular}
\label{tab:obs_ant}
\end{table}

The parameters used for the experiments are a feed-forward neural network with one hidden layer of 100 neurons. The maximum duration of each episode is set to 500 steps with the termination condition of the robot falling down. Regarding the parameters of the algorithm, a total of 100 million evaluation steps were used, and 4 replications to compare statistically the results obtained. 

\section{Results}
This section presents and discusses the results obtained. The results are divided in the behavior with and without neuro-regulation to compare the performance and influence of the technique. 

When testing the version without neuro-regulation, it was found that the robot is not able to learn to perform both behaviors at the same time to some extent. Probably what happened is that the robots start to develop one of the two behaviours and then the evolutionary process tend to consider the robots that are asked to perform that behaviour good and the robots that are asked to perform the other behaviour bad, somewhat independently for their relative ability. Consequently, the evolutionary process get stuck on the attempt to optimise a single behaviour only.

Two ways were found that could alleviate the problem. The first strategy consists in evaluating the robot for two episodes in which they are evaluated for performing the first and the second behaviour. The total fitness will correspond to the sum of the fitness obtained during the two episodes. The second strategy consists in evaluating symmetric individuals for the ability to produce the same type of behaviour. The usage of symmetric samples implies that the centroid is moved in the direction of the individual of the couple that achieved the highest fitness and in the opposite direction of the individual that achieve to lowest fitness. If the two robots are asked to produce different behaviors, the centroid will be moved in the direction of the robot evaluated on the first behaviour, independently for its real ability. If the two robots are evaluated on the same behaviour, things should work better.

\begin{figure}[!h]
    \centering
    \includegraphics[width=\linewidth]{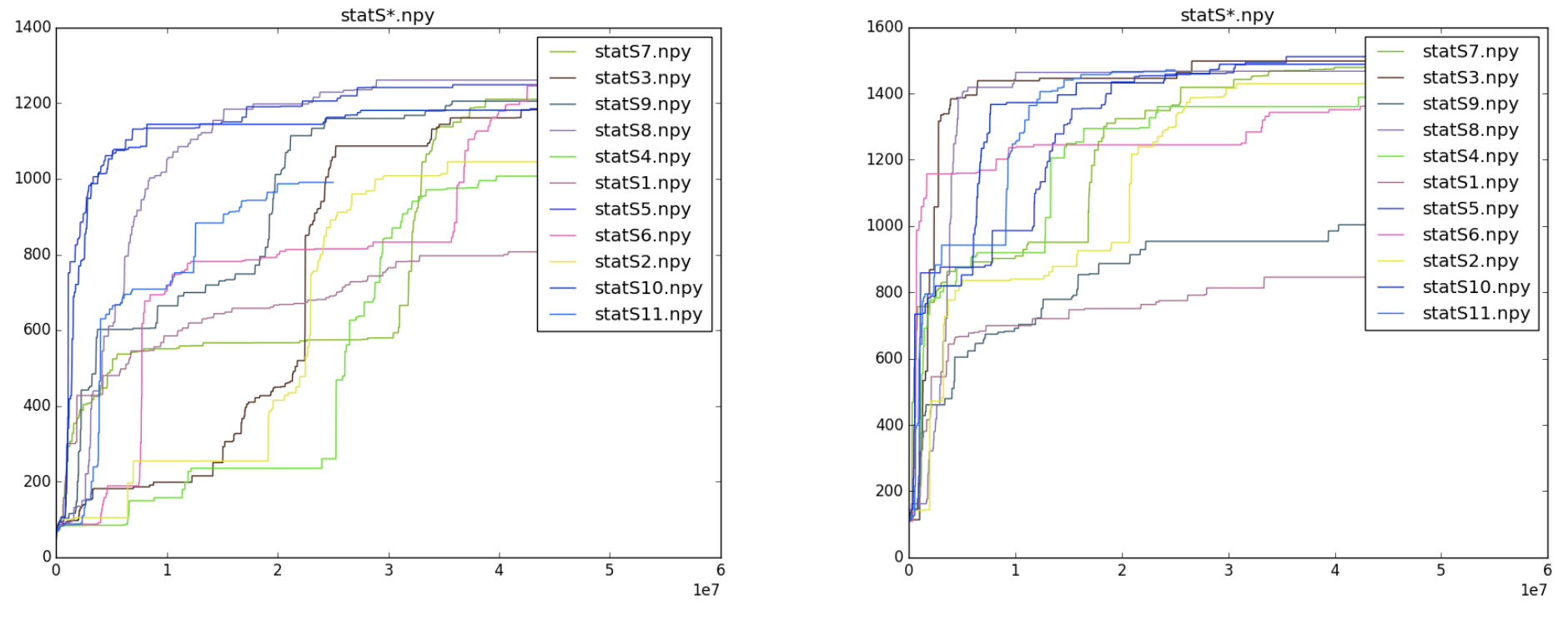}
    \caption{Fitness evolution of the Hopper for 11 replications using strategy 1 (left) and strategy 2 (right), where strategy 1 refers to evaluating the robot for two episodes (one for each behavior) and strategy 2 refers to evaluate symmetric individuals.}
    \label{fig:v1}
\end{figure}

The results showing the comparison of the fitness progress for different seeds is depicted in Figure \ref{fig:v1} for both strategies. Strategy 1 refers to evaluating the robot for two episodes, even episode with the behavior 1 and odd episode with behavior 2. This strategy does not involve randomness, whereas strategy 2 refers to the evaluation of symmetric individuals in the Salimans algorithm, meaning that when an individual needs to be evaluated, a behavior is chosen randomly for the pair. 

\begin{figure}[!h]
    \centering
    \includegraphics[width=0.7\linewidth]{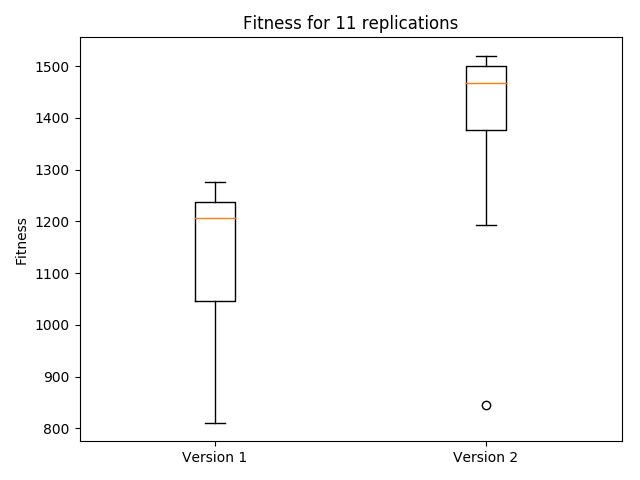}
    \caption{Comparison of the fitness of the Hopper across replications for both strategies with the average represented in orange and its corresponding standard deviation.}
    \label{fig:box}
\end{figure}

Moreover, analyzing Figure \ref{fig:box}, we can conclude that the second strategy performs better than the first one. The first strategy has an average fitness of 1137.19 with a standard deviation of 139.55. On the other hand, the second strategy has an average fitness of 1382.38 with a standard deviation of 192.29. 

\begin{figure}[!h]
    \centering
    \includegraphics[width=\linewidth]{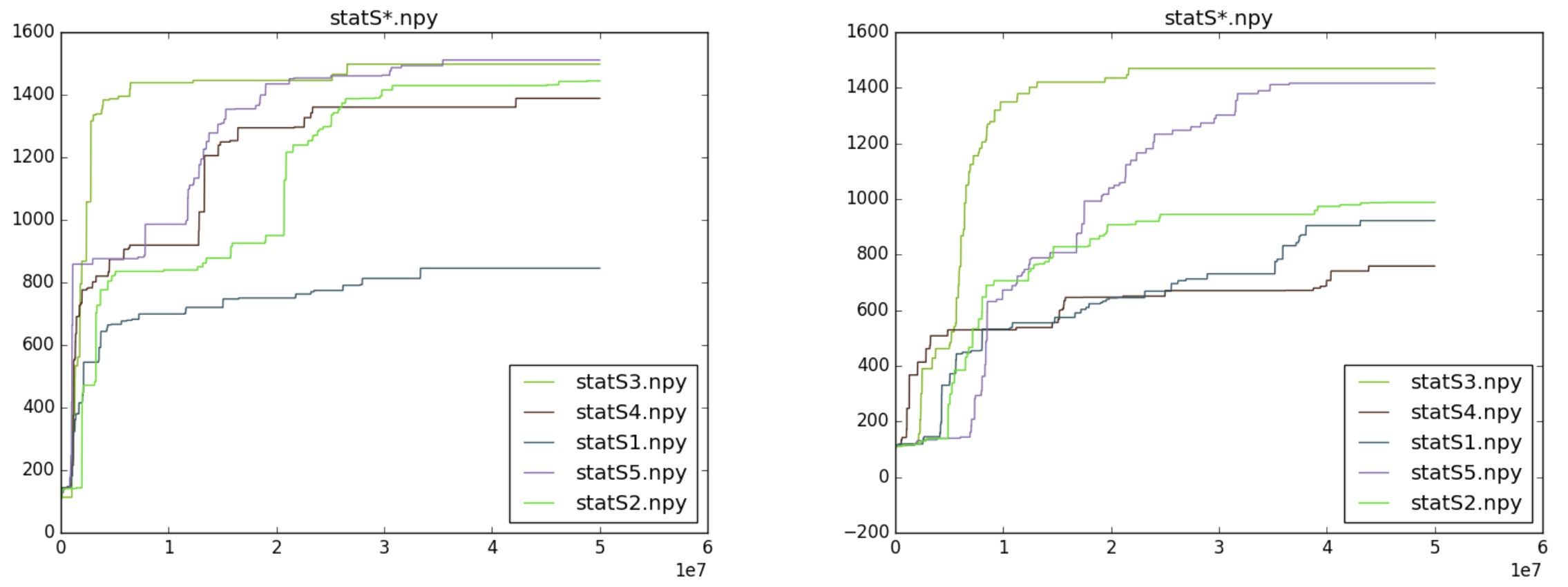}
    \caption{Fitness evolution of the Hopper for 5 replications using strategy 2 for the standard network (left) and neuro-regulation (right)}
    \label{fig:nr-v2}
\end{figure}

\begin{figure}[!h]
    \centering
    \includegraphics[width=0.7\linewidth]{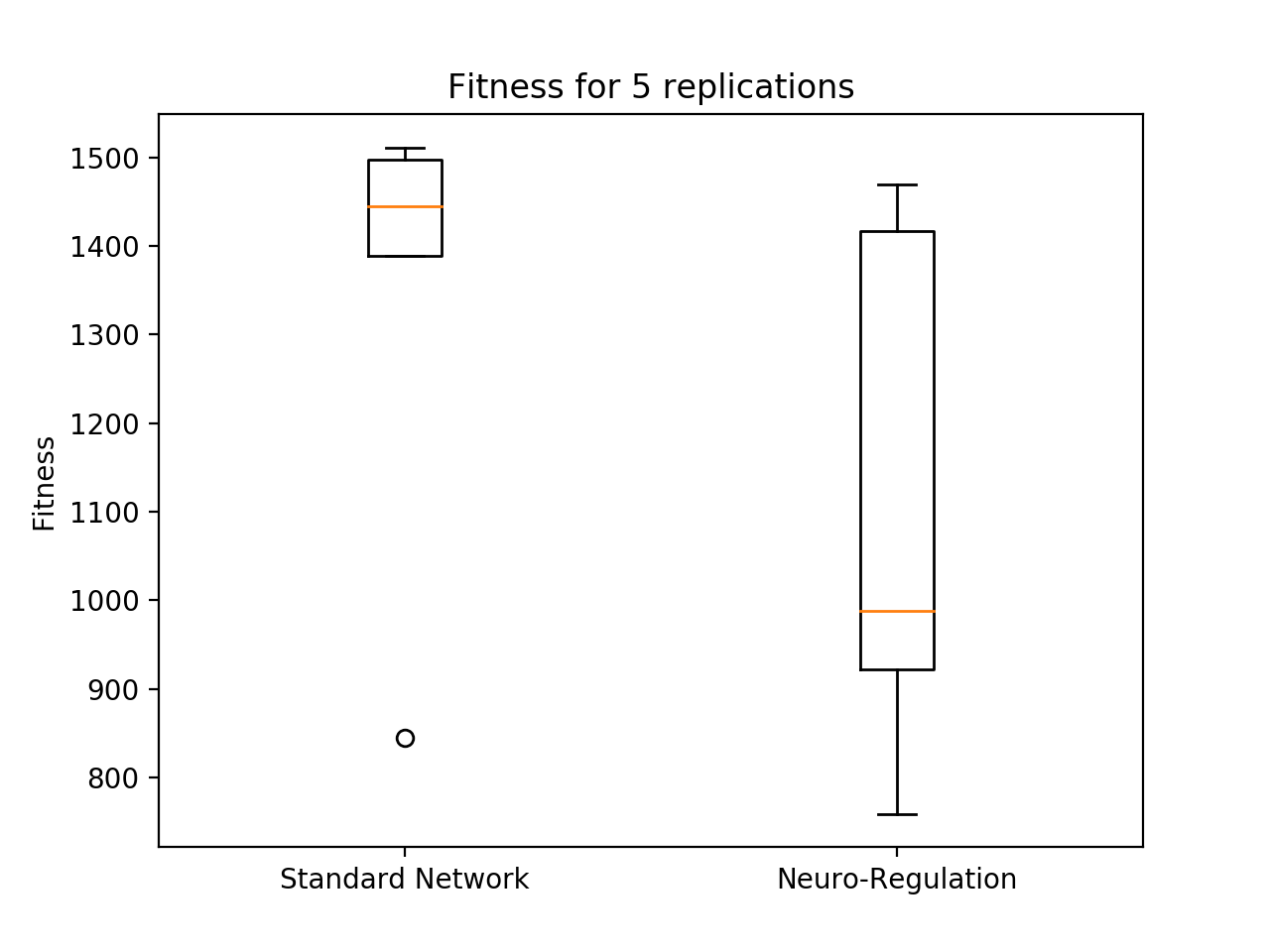}
    \caption{Comparison of the fitness of the Hopper for 5 replications between the standard network with the first strategy (left) and neuro-regulation (right).}
    \label{fig:nr}
\end{figure}

Finally, neuro-regulation is implemented on top of the strategy two, since in principle, it does not influence the strategy chosen to see if it leads to significant improvements. Figure \ref{fig:nr-v2} depicts the evolution of the fitness during five replications for the Hopper. If compared with the raw standard network, there is no significant improvement, even some of the replications perform worst than the standard network. It can be seen in Figure \ref{fig:nr} that there is no big change with the addition of the neuro-regulation. The explanation why modularity helps only a little might be that the performance are already rather good with the standard network. 

\begin{figure}[!h]
    \centering
    \includegraphics[width=\linewidth]{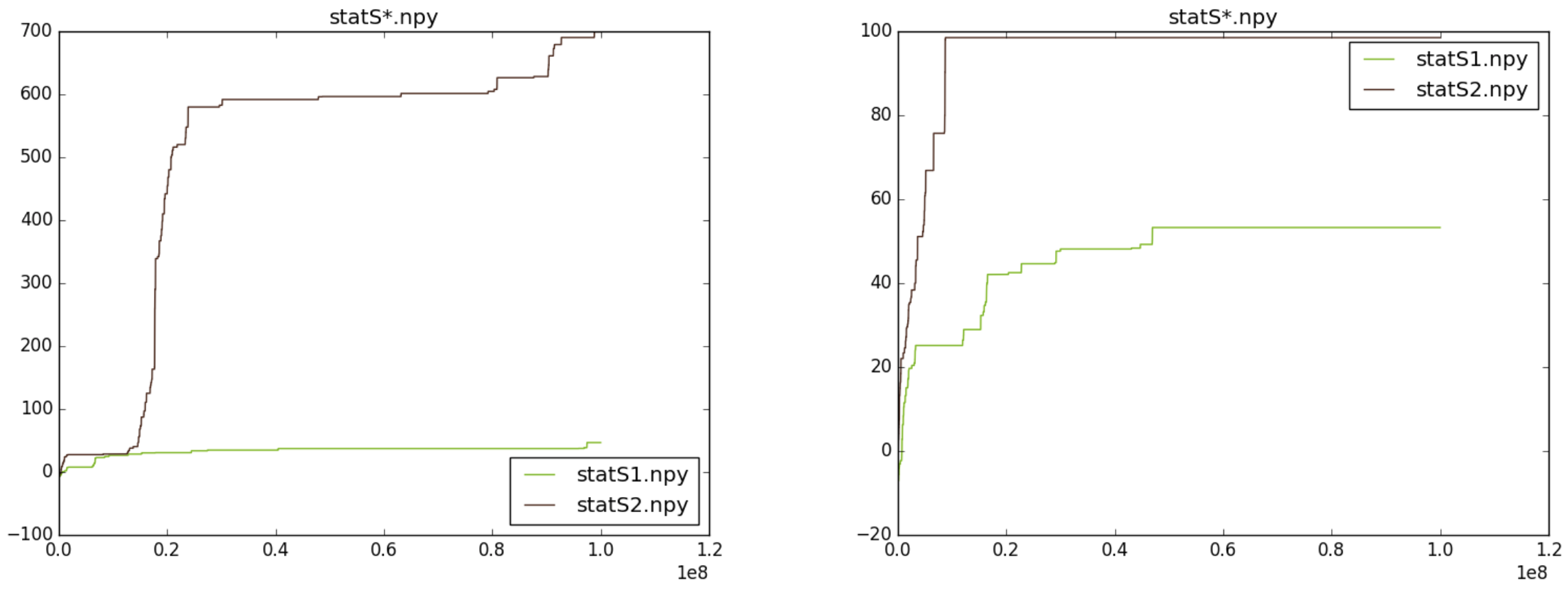}
    \caption{Fitness evolution of the Ant for 2 replications using strategy 2 for the standard network (left) and neuro-regulation (right)}
    \label{fig:ant_stats}
\end{figure}

Regarding the Ant robot, Figure \ref{fig:ant_stats} shows the evolution of the fitness during the training process of 100 million steps using a standard network and neuro-regulation. 

\begin{figure}[!h]
    \centering
    \includegraphics[width=0.7\linewidth]{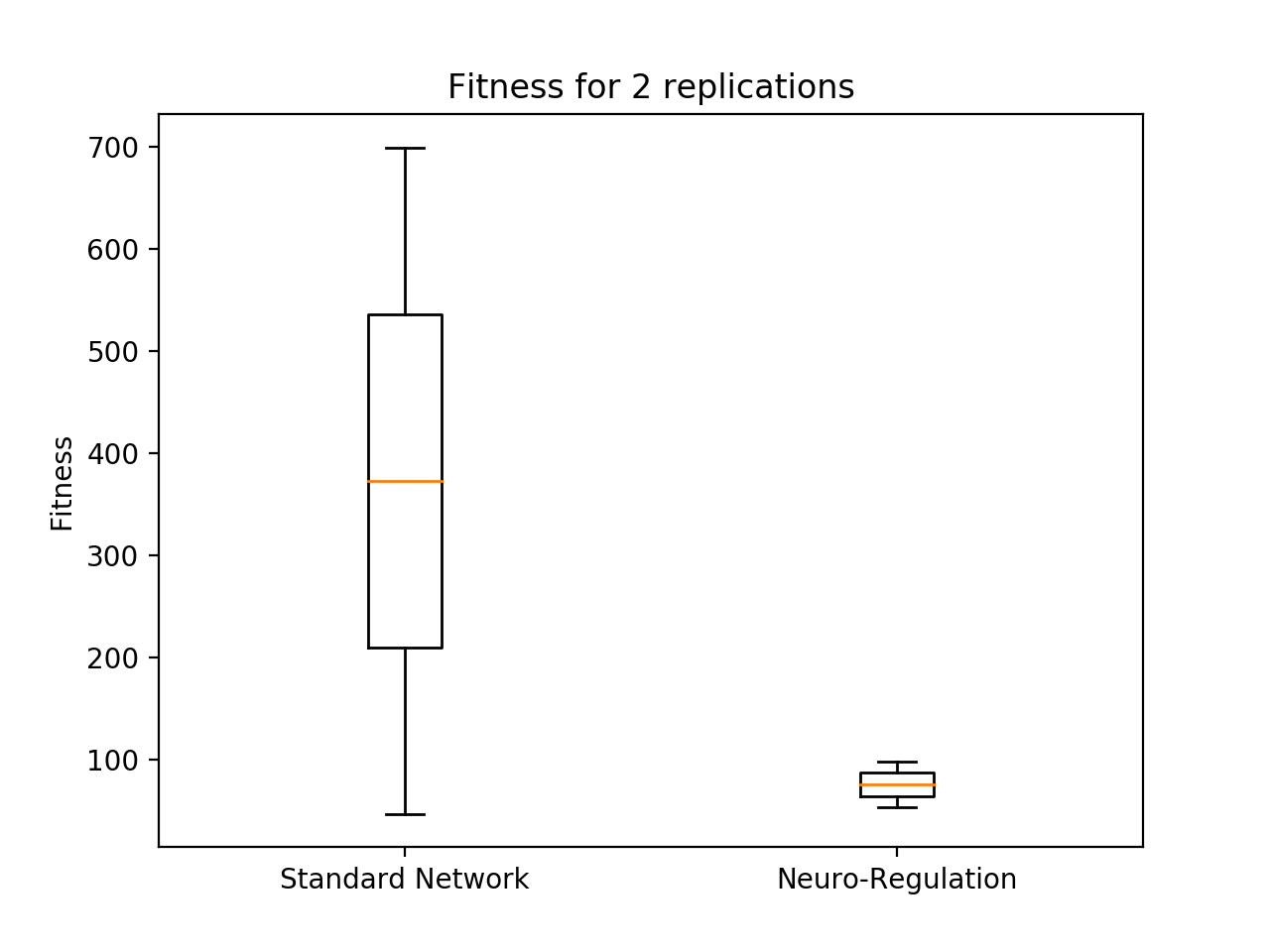}
    \caption{Comparison of the fitness of the Ant for 2 replications between the standard network with the first strategy (left) and neuro-regulation (right).}
    \label{fig:ant_boxplot}
\end{figure}

Moreover, Figure \ref{fig:ant_boxplot} describes the average fitness of the replications set using the standard network and neuro-regulation strategies. 
The standard network seems to obtain better results as compared to the implementation of neuro-regulation. Possibly, more replications are needed to obtain a more robust statistical result. 

The full source code for each step as well as the trained files can be found in the Github repository provided \footnote{\href{https://github.com/vicmassy/behavioral_robotics/tree/master/project}{https://github.com/vicmassy/behavioral\_robotics/tree/master/project}}.

\section{Conclusions}
This project tries to analyze the role of modularity and neuro-regulation for the production of multiple behaviors, by first implementing two strategies for the Hopper robot that learn to perform both behaviors to some extend and then are compared with the addition of neuro-regulation. 

The system does not show significant improvements with the addition of neuro-regulation due to the fact that the robot is already good with the standard network, possibly because of its simplicity. 

In future work more replications must be done regarding neuro-regulation to have a more consistent statistical result to compare with the standard network. Moreover a harder problem must be chosen to see the effects on different environments and robot complexities.

\bibliographystyle{./bibliography/IEEEtran}
\bibliography{references.bib}

\end{document}